\documentclass{article}
\usepackage[preprint]{neurips_2023}

\usepackage[utf8]{inputenc} 
\usepackage[T1]{fontenc}    
\usepackage{hyperref}       
\usepackage{url}            
\usepackage{booktabs}       
\usepackage{amsfonts}       

\usepackage{nicefrac}       
\usepackage{microtype}      
\usepackage{xcolor}         
\usepackage{graphicx}

\title{Context Matters: Data-Efficient Augmentation of Large Language Models for Scientific Applications}

\author{%
    Xiang Li$^{1, 2}$ \quad
    Haoran Tang$^{2}$ \quad
    Siyu Chen$^2$ \\[3pt]
    \textbf{Ziwei Wang}$^2$ \quad
    \textbf{Anurag Maravi}$^2$ \quad
    \textbf{Marcin Abram}$^{1, 3}$\\[4pt]
    $^1$Department of Physics and Astronomy, University of Southern California, Los Angeles, CA, USA\\[2pt]
    $^2$Department of Computer Science, University of Southern California, Los Angeles, CA, USA\\[2pt]
    $^3$Information Sciences Institute, University of Southern California, Los Angeles, CA, USA\\[4pt]
    \texttt{\{limike,haoranta,schen809,zwang476,maravi,mjabram\}@usc.edu}\\
}

\begin{document}

\maketitle

\begin{abstract}
    In this paper, we explore the challenges inherent to Large Language Models (LLMs) like GPT-4, particularly their propensity for hallucinations, logic mistakes, and incorrect conclusions when tasked with answering complex questions. The capacity of LLMs to present erroneous answers in a coherent and semantically rigorous manner further complicates the detection of factual inaccuracies. This issue is especially pronounced in fields that require specialized expertise. Our work delves into these challenges, aiming to enhance the understanding and mitigation of such errors, thereby contributing to the improvement of LLM accuracy and reliability in scientific and other specialized domains. Our findings reveal a non-linear relationship between the context's relevancy and the answers' measured quality.
    In addition, we demonstrate that with the correct calibration, it is possible to automate the grading procedure -- a finding suggesting that, at least to some degree, the LLMs can be used to self-examine the quality of their own performance. Finally, we describe an experimental platform that can be seen as a proof-of-concept of the techniques described in this work.
\end{abstract}

\section{Introduction}

    Large Language Models (LLMs) such as GPT-4 suffer from hallucinations, logic mistakes, and incorrect conclusions when answering challenging questions \citep{rawte2023troubling}.
    In addition, similar to a style adoption \citep{Gatys2016}, LLMs can present erroneous answers in a seemingly coherent and semantically rigorous fashion \citep{Bender2021, Scheurer2023}. Therefore, detecting factual mistakes made by LLMs can be challenging, especially when asking about topics requiring specialized expertise.
    
    We have many available benchmarks measuring the performance of LLMs, including AGIEval \citep{zhong2023agieval}, ChemLLMBench \citep{guo2023chemllmbench}, and SCIEval \citep{sun2023scieval}. However, those popular benchmarks usually require a specific structure of the responses (e.g., multiple choice or numeric) and are not designed to capture mistakes in free-form responses.
    Among more flexible methods, we can mention HaluEval developed by \citet{li2023halueval}. However, this method does not scale well since it requires laborious and expensive human annotation. 
    Automated methods of evaluations include BiLingual Evaluation Understudy (BLEU) \citep{BLEU}, BERTScore \citep{BERTScore}, and LLM-Eval \citep{LLM-eval}. While they can be easily run at scale, they are usually restricted to a group of specific sub-tasks (e.g., machine translation, text summarization, or dialogue generation).
    Due to those constraints, the question of LLMs evaluation is still an active research area.
    
    In many areas, the tendency of LLMs to hallucinate and its typical overconfidence when presenting erroneous answers can cause a lot of harm (e.g., in medical \citep{Thapa2023}, legal \citep{Weidinger2022}, or educational \citep{Bommasani2021} applications). In this work, we focus on yet another area characterized by a low tolerance for erroneous answers: \emph{scientific applications}, with the aim to develop a technique that increases the coherence and reliability of answers produced by popular LLMs, such as GPT-3.5 and GPT-4.

    We tackle this challenge by employing a version of prompt engineering known as \emph{the chain-of-thought} \citep{wei2023chainofthought}. As one of the original contributions of this paper, we measure the extent to which the relevance of context matters in improving the quality of answers when asking about topics requiring specialized expertise. Next, in Section \ref{sec:results}, we discuss to which extent we can automate the evaluation procedure, testing two distinct approaches: (1) we anchor the ground truth by providing a sample answer, (2) we anchor on the generated answer based on the sample answer as a fact.
    
    To illustrate our approach, we developed an online platform that can serve as a proof-of-concept, testing the techniques described in this paper. As a choice of our scientific domain, we selected graduate-level quantum physics. In Section \ref{sec:proof_of_concept} we discuss the platform's architecture
    and in Section \ref{sec:summarization}, we test the impact of different summarization techniques for keeping the context reasonably short.

\section{Characteristic Challenges and Research Opportunities}  

    Arguably, the fastest adoption of LLMs in the past year could be noticed in the areas of translation \citep{xu2023paradigm}, code generation \citep{koziolek2023llm, siddiq2023generate, chen2023effectiveness}, database query optimization \citep{Li2023}, and text summarization \citep{sahu2023enchancing} as well as in the development of general personal assistants \citep{li2023human, chan2023mango}. 
    
    However, when targeting \emph{scientific applications}, we must be aware of some distinctive characteristics of that domain that set it apart from typical business operations. For example, when answering scientific questions, we might face some inherent limit on the size (or variability) of the known datasets. While in many business applications, we can get more trainable examples by committing more money or time (e.g., by taking more photos and labeling them or by scanning more books to get more natural text), in some scientific domains, the number of existing examples can be strictly limited. For example, we might only know so many examples of superconducting materials, and we can not easily increase that number (not without first discovering new types of superconductors).
    Another important difference is that science does not follow the typical majority voting rule. If we ask people to classify pictures of cats and dogs, even if some human labelers might disagree when assessing a particular photo, we can still use majority voting to determine whether a given picture depicts a dog or a cat. The same can be done to quantify the value of continuous variables \citep{Surowiecki2004}. In contrast, in science, a single opinion, if sufficiently motivated, can weigh more than a popular opinion of the majority (as it happened, e.g., with the discovery of quasicrystals by \citep{Shechtman1984}, a discovery that challenged the existing status quo, and resulted in a major shift in our understanding of solid-state physics).
    On the other hand, in the scientific domain, we can rely on some known (even if sometimes only in an approximate form\footnote{
        For example, Newton's third law is valid in classical mechanics but invalid in electrodynamics \citep{Taylor2005} -- though the deviation from Newton's third law is significant only for strong magnetic fields or when relativistic effects are significant. Another example can be the conservation of mass, which is valid only for closed systems \citep{Xia1992}.})
    fundamental laws governing the universe (e.g., the laws of motion or the conservation of energy, momentum, angular momentum, charge, mass, etc.). This contrasts with many business applications that often target complex social systems, where emerging phenomena render any fundamental description unusable \citep{Anderson1972}. For example, when predicting the probability that a given customer will return in the next six months, we can not rely on any strict laws we can use to verify the correctness of our solutions. This makes applications in the scientific domain somehow special since we can verify with certainty that answers violating some fundamental laws are wrong.
    
    Considering all those differences, we must leverage the few-shot learning ability of LLMs \citep{brown2020language}.
    The algorithm that we propose must be data-efficient. We can not afford to retrain the entire LLM using millions of examples, as simply they might not exist. We must leverage the few-shot learning ability of LLMs. Instead, we focus on providing LLMs with the proper context, as described in the following sections.
    
\section{Related Work}

    \paragraph{Prompt engineering.}  Previous work has demonstrated that prompt engineering and chain of thought techniques can greatly improve LLMs' performance in various domains \citep{kojima2023large,wei2023chainofthought, besta2023graph,yao2023tree,imani2023mathprompter}. However, very few works focus on the evaluation of these techniques when applied to the scientific domain.

    \paragraph{In-Context Learning.} The few-shot learning capabilities of LLMs enable enhanced performance across various tasks through tailored prompting \citep{brown2020language}. \citet{Drori_2022} employed OpenAI's Codex to tackle University Math Problems, achieving human-level performance and elevating benchmark accuracy from $18.8\%$ to $81.1\%$. 
    \citet{imani2023mathprompter} developed a ``math-prompter'' method for LLMs, significantly improving performance on the MultiArith dataset from $78.7\%$ to $92.5\%$. \citet{zhang2023makes} developed a concise two-stage framework that first generates rationales and then uses them to enhance answer inference, effectively integrating multimodal information. However, today's state-of-the-art LLMs (e.g., GPT-4) still struggle to solve advanced, graduate-level physics problems, as demonstrated in this paper. Key issues include its poor handling of physical units, coding errors, and its tendency to create convincing yet incorrect answers. 

    \paragraph{Evaluation of LLMs' physics reasoning capacities.} Creating an effective evaluation framework for LLMs in scientific domains, particularly for physical reasoning, is challenging due to the implicit nature of human knowledge about the physical world and its representation \citep{bisk2019piqa}. Benchmarks like those created by \citet{clark2018think} have focused on assessing LLMs using physics questions. \citet{arocaouellette2021prost} specifically test models on physical commonsense and reasoning concepts. \citet{liang2023code, ahn2022i} focus applications for physically grounded contexts. However, most of those benchmarks target basic school- or undergraduate-college-level questions. 
    In contrast, in this work, we focus on the topic of upper-level undergraduate- and graduate-level quantum physics to provide LLM with more nuanced questions that often do not have a single, straight answer.

\section{Methodology}

\subsection{Approach}

    To assess the importance of context, we compare the answers of GPT-4 in two settings: (1) without any modifications, a version that can be seen as zero-shot learning, and (2) using prompt engineering to provide the context. We vary the quality of the context, from ``irrelevant'', through ``vague'' to ``insightful''. We score the quality of the response using developed grading rubrics. In addition, we test a few variants of automatic grading, measuring its consistency with the manual expert grading. Finally, we test the impact of summarization techniques employed to limit the length of the context.
    

\subsection{Data collection}

    We created a question bank containing selected problems from \citep{Griffiths2018}, a popular upper-undergraduate or intro-graduate-level textbook on quantum mechanics. Next, for each question, we created three hints with different levels of relevance.
    
    The selected questions cover the following areas: quantum mechanics in three dimensions, identical particles, symmetries and conservation laws, time-independent perturbation theory, and the variational principle (cf. the sample question in the Supplementary Material). The solutions to the questions usually involve a combination of complex calculations and application of conceptual knowledge.

\subsection{Hints structure}

    Hints for each question are divided into three classes: \emph{irrelevant hint}, \emph{vague hint}, and \emph{insightful hint}. Insightful hints are hints that contain information highly relevant to the question. These hints can contain a sample solution for a similar problem, key insights into the formation of the question, or methodology that can be applied to the answer. Vague hints are hints relevant to the topic of the question but not the question itself. For example, if the question asks about the variational principle, a vague hint can be the sample solution to a completely different variational principle question or a textbook description of the variational principle. Unrelated hints include the solution to a difficult quantum physics problem, which is not relevant to the topic of the question.

    You can find examples of hints representing each category in the Supplementary Material.

\subsection{Evaluation}

    To assess different aspects of answers produced by the LLM, we developed grading rubrics evaluating not only the semantic accuracy of the responses but also the factual correctness and their logical consistency. Namely, our evaluation system comprised:
    \begin{itemize}
      \item \textbf{Completness and Relevancy (3 points):} Three points for a relevant and correct response. Zero points if the answer is irrelevant or incorrect, with partial points for partial correctness.
      \item \textbf{Logic and Reasoning (4 points):} We start with four points and deduct two points for each logical inconsistency and one point for each minor error in reasoning.
      \item \textbf{Truthfulness (lack of hallucination, 3 points):} We start with three points and deduct one point for each incorrect statement in the response.
    \end{itemize}

    We graded each response using both manual and automatic approaches. As for manual grading {\tt (manual)}, we shuffled the order of responses to reduce bias (so the grader does not know which response is the result of which). The grader then used the evaluation criteria and sample answers to grade. For automatic grading, we have implemented two approaches. The first approach {\tt (auto1)} involved providing GPT-4 with an answer and rubric while prompting GPT-4 to grade in a zero-shot manner. In the second approach {\tt (auto2)}, we first prompted GPT-4 to generate an answer based on the sample answer and then used its answer as the key to grade the response. In this way, the grader kept a better version of its response as the benchmark (assuming the quality of the response is directly and positively correlated to the relevance and correctness of the prompt).

    Since we have four levels of context (``no context'', ``insightful hint'', ``vague hint'', and ``irrelevant hint''), a total of four responses were generated by GPT-4 for each question.
    You can find a sample of grading with our evaluation criteria in Supplementary Material.
    
\section{Results}
\label{sec:results}

\subsection{Context matters}

    We provide GPT-4 with different levels of context when answering the queries from our question bank. The context is provided to GPT-4 API with an ``assistant'' role, followed by a ``user'' call to answer the question. For each question, we test the response, providing four different levels of context. In the first level (no hint), we provide no context. In the second level (irrelevant hint), the context contains a hint that is relevant to quantum physics but not relevant to the topic of the question. In the third level (vague hint), the context contains a hint that is relevant to the topic of the question but not directly applicable or related to the question. In the fourth level (insightful hint), the context contains a relevant and applicable hint to the solution.

\begin{table}[b]
    \centering
    \caption{
        Mean grading score for different levels of hints and different grading modes (manual and two different auto-grading, as explained in Section \ref{sec:auto-grad}). The error represents the unbiased sample standard deviation of the mean values.
    }
    \label{table:1}
    \begin{tabular}{l c c c c} 
     \toprule
        & No hint & Irrelevant & Vague & Insightful \\
     \midrule
         Manual & $7.5 \pm 0.6$ & $7.9\pm 0.6$ & $7.9\pm 0.7$ & ${\bf 8.4 \pm 0.6}$ \\
         Auto1 & $6.6\pm 0.6$ & $6.8\pm 0.4$ & $6.6\pm 0.7$ & ${\bf 7.4\pm 0.6}$ \\
         Auto2 & $6.4\pm 0.6$ & $7.0\pm 0.4$ & $6.6\pm 0.7$ & ${\bf 7.6\pm 0.5}$ \\
     \bottomrule
    \end{tabular}
\end{table}

    After acquiring the responses from GPT-4, we shuffled them and manually graded them according to the evaluation criteria. We present the mean scores in Table \ref{table:1} (where we also included results of auto-graders, explained below). The results suggest that different levels of quality of the context leads to different performance. Notably, even irrelevant context seems to improve the average performance.
    
\subsection{Automatic grading}
\label{sec:auto-grad}

\begin{figure}
    \centering
    \includegraphics[width=0.95\textwidth]{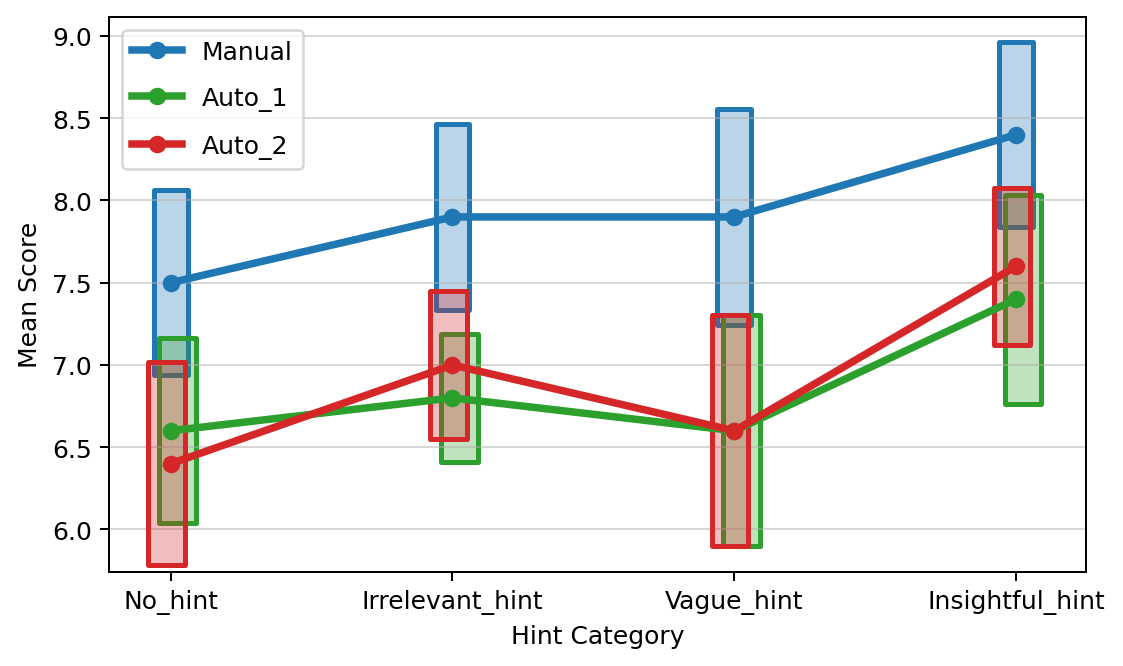}
    \caption{Mean grading score for different levels of hints and different grading modes. The error boxes represent the unbiased sample standard deviation of the mean values. While there is no statistical difference between the two tested auto-grade methods, the manual method yields higher scores than the automated method.} 
    \label{fig:trend}
\end{figure}

    Aside from grading manually, we also experimented with three versions of automatic grading with GPT-4. The first version (auto-1) grades by treating the sample answer as the ground truth, and the second version (auto-2) first generates an answer based on the sample answer provided as a context, before using that generated answer as the ground truth. We present the averaged results in Table \ref{table:1} and Figure \ref{fig:trend}.
    
    Analyzing the general trend depicted in Figure \ref{fig:trend}, both auto1 and auto2 grading methods yield statistically similar results but differ from the manual grading. This gap between manual and automated scores can be explained by a lack of calibration of the automatic scoring system. Nevertheless, the fact that the general slope of each line is similar might suggest that automatic grading can be a viable alternative to manual grading, allowing us to scale our methodology in the future. 

\begin{figure}
    \centering
    \includegraphics[width=\textwidth]{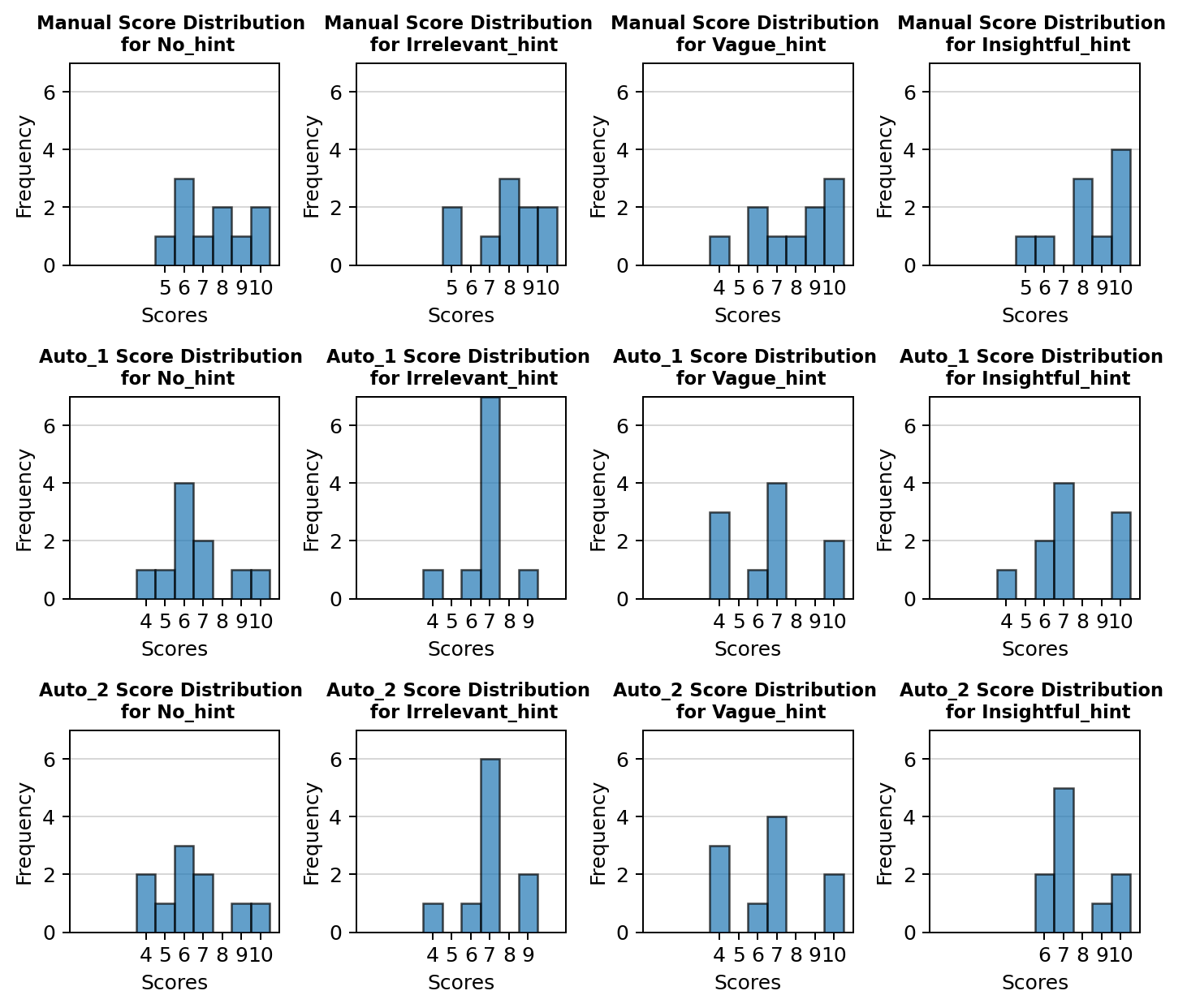}
    \caption{Score Distribution for responses generated with different prompts and grading techniques} 
\label{fig:score}
\end{figure}

    In Figure \ref{fig:score}, we present detailed histograms of scores for each question and hint type. In Table \ref{table:2}, we depicted the population standard deviation of each score group. Notably, we obtain the highest standard deviation for the vague-hint category. This suggests that while low-quality context can improve the mean performance, it also increases the variability of responses.

\begin{table}[b]
    \centering
    \caption{
        The measured sample standard deviation of each score group.
    }
    \label{table:2}
    \begin{tabular}{l c c c c}
     \toprule
        & No\_hint & Irrelevant & Vague & Insightful\\
     \midrule
         Manual & 1.69 & 1.70 & {\bf 1.97} & 1.69\\
         Auto1 & 1.69 & 1.17 & {\bf 2.11} & 1.91\\
         Auto2 & 1.85 & 1.34 & {\bf 2.11} & 1.43\\
     \bottomrule
    \end{tabular}
\end{table}
    
\subsection{Consensus between manual grading and automatic grading}

    Consensus across all three grading strategies highlighted two key insights: firstly, insightful hints notably enhance GPT-4's performance in terms of mean scores; secondly, responses generated without hints performed the worst. 
    
    These results corroborate our initial hypothesis that highly relevant hints significantly boost GPT-4's efficiency in addressing questions of scientific domains. Interestingly, our results also suggest that ``irrelevant hints'' boost the performance of our LLM in a similar way as ``vague hints'' (cf.\ again Figure \ref{fig:trend}). This unexpected outcome could imply a misalignment between human perception of hint relevance and the model's interpretation. It raises the possibility that what seems relevant and helpful to a human may not necessarily resonate in the same way with GPT-4, underscoring the nuanced dynamics of human-AI interaction in educational contexts.

\section{Proof of Concept}
\label{sec:proof_of_concept}

\subsection{QuantumGPT platform} \label{QuantumGPT}

\begin{figure}
    \centering
    \includegraphics[width=0.85\textwidth]{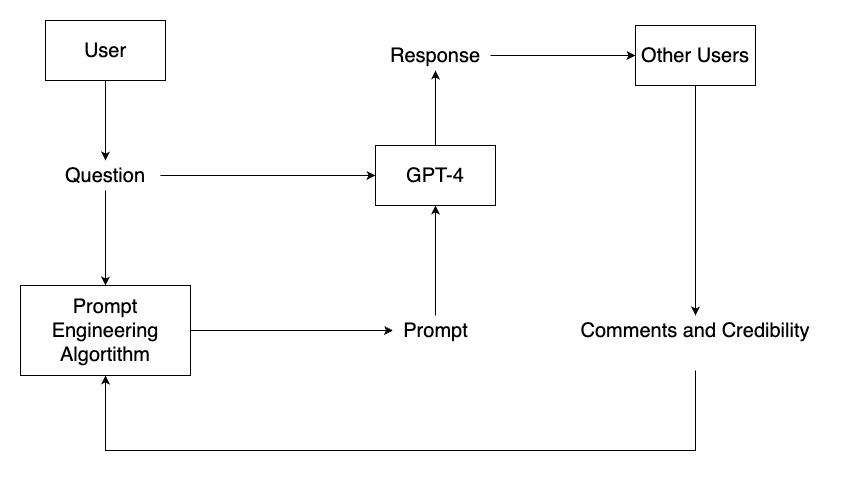}
    \caption{The depiction of the QuantumGPT's pipeline.} 
    \label{fig:overview}
\end{figure}

    To further improve the performance and versatility of our prompt engineering approach, we have designed a platform (denoted below as QuantumGPT and available at \url{www.quantumgpt.science}) that can help us gather more data in a crowd-sourcing manner.
    The general pipeline of QuantumGPT is shown in Figure \ref{fig:overview}. It is a prompt-engineering project that utilizes user feedback to improve the response of GPT-4. In addition, the platform has interactive features, including chat, discussions, and the ability to post blog-style articles. 
    
\subsection{The Credibility-Enhanced Feedback Architecture, CEFA} \label{CEFA_architecture}

\begin{figure}
    \centering
    \includegraphics[width=0.85\textwidth]{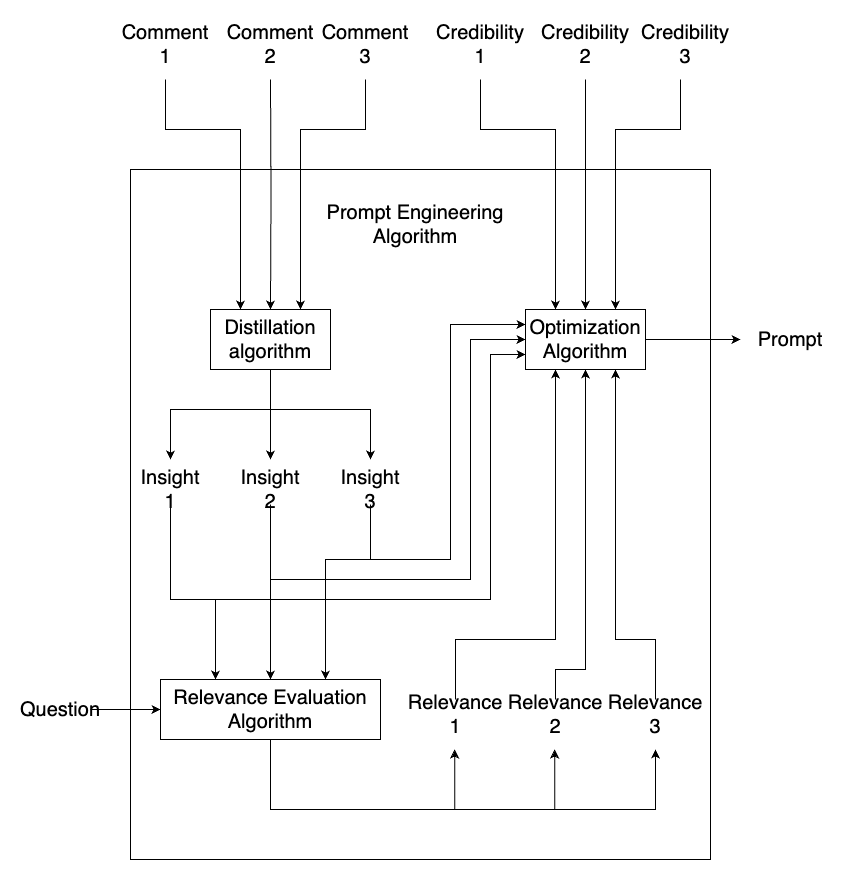}
    \caption{The schematics of the CEFA architecture.} 
    \label{fig:pipeline}
\end{figure}

    The accuracy and quality of data are critical to the performance. Some previous works have discussed the threat of data poisoning attacks \citep{chan2020poison, fowl2021adversarial}. Even without a malicious attack, we are still vulnerable to unintentional inaccurate information. To address this concern and improve the data quality collected, we propose a Credibility-Enhanced Feedback Architecture (CEFA), as outlined in Figure \ref{fig:pipeline}. It consists of three main modules: the distillation algorithm, relevance evaluation algorithm, and optimization algorithm. There is also a separate credibility system to examine the interaction between users and evaluate the credibility of each user. The distillation algorithm distills insights from users' comments. Insights are then passed to the relevance evaluation algorithm to generate a relevance score. Finally, insights with their respective relevance and credibility scores are passed to the optimization algorithm to engineer the prompt.

    \paragraph{Distillation Algorithm} 

    The Distillation Algorithm module is developed to optimize the utility of user-generated content within QuantumGPT, aiming to refine the information flow by reducing the token count of user comments while preserving their intrinsic insights. This algorithm is adept at processing raw user comments, transforming them into a distilled format that retains the core insights but minimizes unnecessary verbosity and redundancy. Such a transformation is instrumental in significantly enhancing the data handling efficiency within the QuantumGPT framework.

    The deployment of these summarized insights contributes notably to the overall performance of QuantumGPT. By reducing the length of the inputs, the algorithm effectively diminishes the computational burden on the system. This reduction in computational load directly translates into an enhanced speed of response generation, thereby elevating the system's capability to handle complex user interactions swiftly and accurately.

    \paragraph{Relevance Evaluation Algorithm}

    The Relevance Evaluation Algorithm is designed to ensure precise and relevant dialogue in quantum physics. It utilizes advanced Natural Language Processing techniques to assess the semantic relevance between user queries and responses, going beyond traditional keyword matching to include semantic similarity and contextual embedding models.

    The algorithm employs sentence-transformer models to generate embeddings from user inputs, analyzing semantic relationships through cosine similarity measures. This process allows for the generation of relevance scores, which indicate the semantic alignment between questions and comments, ensuring contextually relevant and precise responses from QuantumGPT.

    \paragraph{Credibility Algorithm}

    The Credibility Algorithm is designed to enhance online discourse quality by assessing user interactions. This algorithm identifies three ways for users to gain credibility: prior credentials (such as a relevant university affiliation), experience using the platform, and interactions with other users.
    
\begin{figure}
    \centering
    \includegraphics[width=0.8\textwidth]{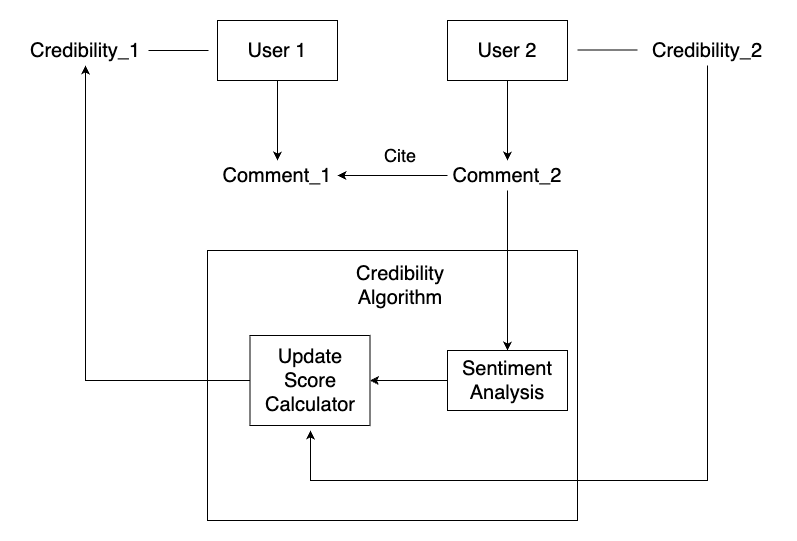}
    \caption{The depiction of the Credibility System.} 
    \label{fig:credibility}
\end{figure}

    The algorithm operates within a dynamic feedback mechanism, as depicted in the Figure \ref{fig:credibility}. It functions by analyzing the sentiment of a comment from User 1 towards User 2. If User 1, who holds a high credibility standing, expresses a positive sentiment in their comment, the algorithm activates the Update Score Calculator. This calculator, in turn, increases User 2's credibility score, provided the sentiment analysis results are positive.

    For the purpose of the sentiment analysis, we use the ``distilbert-base-uncased-finetuned-sst-2-english'' language model. This model, a streamlined version of the BERT framework, has been fine-tuned on the Stanford Sentiment Treebank (SST-2) dataset, making it highly effective in analyzing user comments. The sentiment scores derived from this analysis are used to adjust the credibility scores of users, with a particular emphasis on positive feedback from highly credible users. This process ensures that the platform's credibility ratings reflect constructive and positive user interactions.

\section{Context Window and Summarization Methods}
\label{sec:summarization}

    Each LLM, including GPT-4, can only accept a finite-size input shorter than a specific context window. Therefore, in any practical implementation, we must implement a summarization technique to keep the context concise. In this analysis, we test the impact of the context summarization on the performance of the LLM.
    
    An additional advantage of context summarization is that shorter input reduces the computational load on the system and increases the speed of response generation. For our preliminary methodology, as our primary summarization model, we utilize BART, which has been fine-tuned on the CNN/Daily Mail dataset. This model is not explicitly trained on scientific texts –- a domain that is relatively underrepresented in text summarization models. Despite this limitation, the model demonstrates proficiency in rendering concise summaries.

    Our experiment entailed utilizing the BART model on three user hints ranked in relevancy to each question. We measured the outcomes by assessing the extent of length reduction and the effectiveness in preserving critical concepts.
    For the purpose of this assessment, we modified our auto-grading technique by incorporating the manual chain of thought technique \citep{manualcot} {\tt (auto3)}.  This adaptation involves including our manual grading of a response as an example within the auto-grader system. This modification aims to align the results of the auto grader more closely with the outcomes of manual grading. This approach is expected to enhance the accuracy and reliability of {\tt (auto3)} by incorporating human-like reasoning processes into its grading methodology.

\begin{table}[b]
    \caption{Hints length and summarized insights length (in the number of characters).}
    \label{table:3}
    \centering
    \begin{tabular}{r r r r r r r} 
     \toprule
      & hint1 & insight1 & hint2 & insight2 & hint3 & insight3\\ 
     \midrule
     1 & 168 & 167 & 412 & 275 & 1284 & 286 \\ 
     2 & 412 & 275 & 168 & 167 & 1284 & 286\\
     3 & 68 & 56 & 80 & 68 & 1284 & 286\\
     4 & 80 & 68 & 68 & 56 & 1284 & 286\\
     5 & 70 & 69 & 1687 & 234 & 1284 & 286\\ 
     6 & 1669 & 235 & 552 & 174 & 1284 & 286 \\ 
     7 & 1852 & 211 & 927 & 248 & 1284 & 286 \\ 
     8 & 2106 & 203 & 927 & 248 & 1284 & 286 \\ 
     9 & 1849 & 302 & 1288 & 194 & 1284 & 286 \\ 
     10 & 1617 & 302 & 1288 & 194 & 1284 & 286 \\
     \bottomrule
    \end{tabular}
\end{table}

    \begin{figure}
        \centering
        \includegraphics[width=\textwidth]{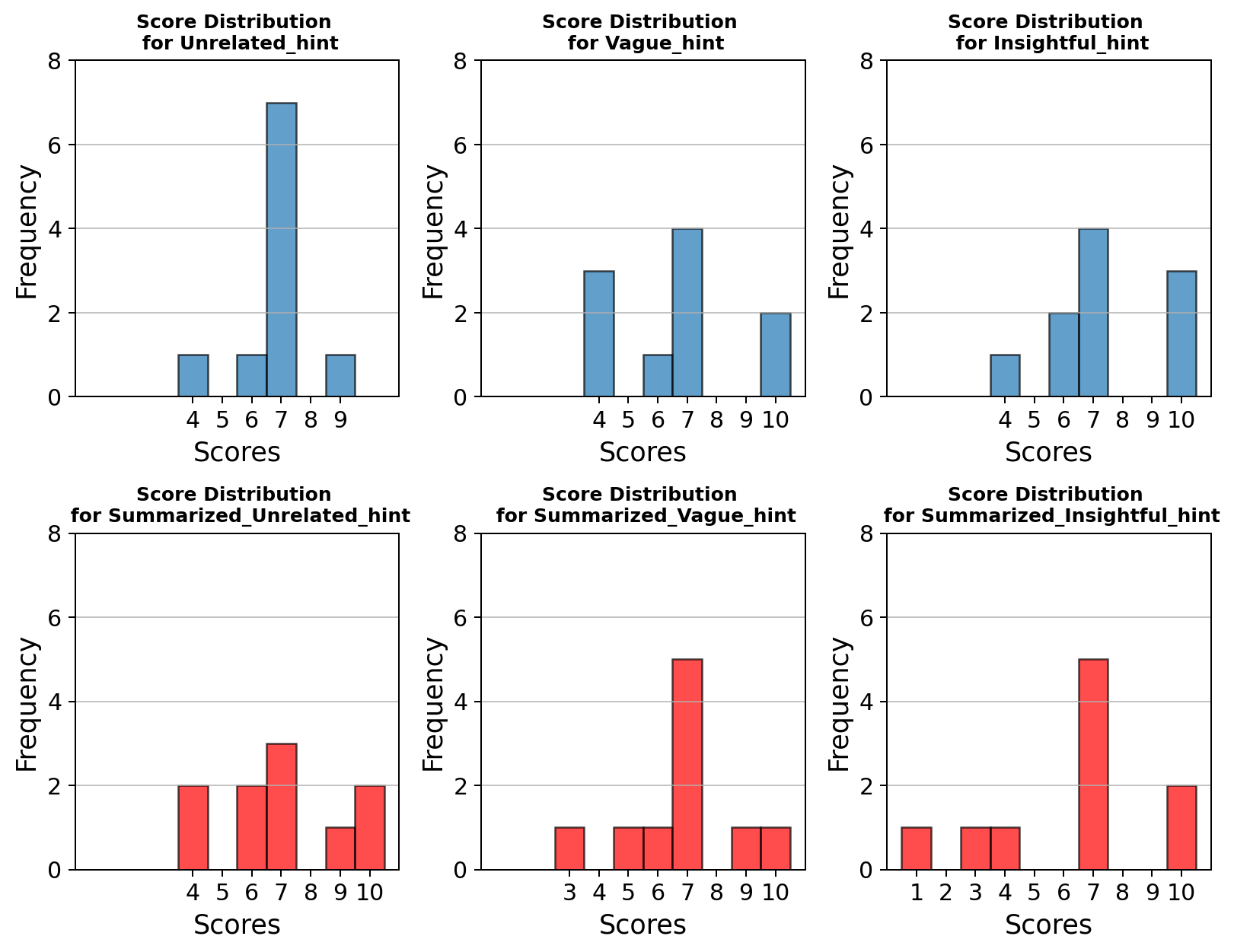} 
        \caption{{\tt Auto3} score distributions for responses generated with access to different level of context, with and without applying the summarization technique.} 
    \label{fig:insight_score}
    \end{figure}

    In Table \ref{table:3}, we show the degree to which we were able to shorten each hint.
    Next, by comparing the correctness of responses prompted with original comments and summarized insights, we measured the quality and relevance of the insights. 
    We show the detailed scores for each type of summarized group of hints in Figure \ref{fig:insight_score}. 

    \begin{table}
    \centering
    \caption{
        Mean Score for Different Summarized Hints and Grading
    }
    \label{table:4}
    \begin{tabular}{l c c c c}
     \toprule
         & Irrelevant & Vague & Insightful\\
     \midrule
         Auto2  & 7.0 & 6.6 & {\bf 7.6} \\
         Auto3+summaries  & {\bf 7.0} & 6.8 & 6.3\\
     \bottomrule
    \end{tabular}
    \end{table}


    In Table \ref{table:4}, we compare the average score on the unabridged and summarized context. We see no reduction in the performance in the cases of the ``irrelevant'' and ``vague'' hints. This might imply that despite the length reduction, the integrity and informative value of the original content were preserved in the summaries.
    However, in the case of the ``insightful'' hints, the reduction in the performance is noticeable. This calls for further studies.
    One way of obtaining better results could be fine-tuning the BART model for scientific domain data -- such intervention should increase the quality of the summaries and might reduce the problem.

\section{Discussion}

\subsection{Significance of Context in LLM Performance}

    Our findings underscore the critical role of context in enhancing the performance of Large Language Models (LLMs) like GPT-4, particularly in the specialized domain of quantum physics. The study reveals that providing contextually relevant hints can significantly improve the accuracy and robustness of LLM responses. However, our result also shows that hints relevant to the topic do not provide any more performance boost when compared to irrelevant hints. One hypothesis is that the determining factor to the helpfulness of the hint is deep insights into the methodology relevant to the question, and insights into the methodology can exist in hints that are not relevant to the topic. Currently, we are working on expanding our dataset to increase the statistical significance of our results.

\subsection{Promise of Automatic Grading}

    The exploration of automatic grading methods presents great opportunities. Our experiments with auto1 and auto2 grading methods align decently well with manual grading outcomes (at least, in terms of the perceived sensitivity of this method). This suggests the potential of these methods as cost-effective and scalable alternatives to manual grading. However, the observed discrepancies, particularly in assessing unrelated and vague hints, highlight the need for further refinement and understanding of how LLMs interpret and respond to different types of context.

\subsection{Challenges and limitations}

    Although our preliminary results are interesting, we are aware of the small size of our data set (10 entries and 40 responses in total). It should be pointed out that creating hints and conducting manual evaluation is a difficult task that requires domain knowledge. We are currently working on expanding the dataset and collecting questions and feedback regarding our experimental platform, QuantumGPT.

    Additionally, while using BART, fine-tuned on the CNN/Daily Mail dataset, is effective for condensing hints into insightful summaries, we have observed a notable loss of critical information, including equations and scientific notations. Given the scarcity of models trained specifically on scientific texts, there is a compelling case for developing a specialized model. Such a model should be adept at summarizing scientific content, striking a balance between brevity and the retention of rich contextual information. This approach would address the gap in processing and summarizing complex scientific sentences.

\subsection{Future Work}

    We have identified three pivotal areas for future research. Firstly, the investigation of insight distillation methods. This involves refining algorithms to more effectively condense complex scientific information into concise yet comprehensive insights, enhancing the efficiency and accuracy of LLM responses. Secondly, the further development of automatic grading systems. Lastly, an experiment with the CEFA architecture to test the users' incentives. Another area of interest is the system's robustness, e.g., its sensitivity to noise or inaccurate information.

\bibliographystyle{plainnat}
\bibliography{bibliography}

\section*{Data and Code Availability}

    Data and code can be found in the following GitHub repository:
    \url{https://github.com/mikelixiang88/Automatic_gpt_grader.git}

\section*{Acknowledgements}
    We would like to take this opportunity to thank Dr. Stephan Haas for providing critical suggestions in designing the research. We would also like to thank Dr. Daniel Lidar for providing challenging questions and feedback on GPT4's performance in the realm of quantum computing. We would like to thank the users of QuantumGPT for providing feedback and data.

\section*{Author Contributions}

    X.L. constructed the quantum question bank, developed the evaluation and auto-grading part of the paper, and proposed the CEFA architecture. H.T. developed the context summarization methods part and part of the auto-grading module, S.C. and A.M. implemented the Relevance Evaluation algorithm, A.M. worked on the optimization algorithm, Z.W. worked on the credibility system and implemented the sentiment analysis module, M.A. provided supervision, proposed the experiment measuring impact of the context, and contributed to the evaluation and the auto-grading part of the paper. All the authors contributed to writing the original draft of this article.

\section*{Competing Interests}

    The authors declare no competing interests.

\clearpage
\appendix

\section*{\Large Supplementary Material}

\section{Sample question}
"Question": "At time t = 0 a large ensemble of spin-1/2 particles is prepared,
all of them in the spin-up state (with respect to the z-axis). They are not 
subject to any forces or torques. At time t1 > 0 each spin is measured--some
along the z direction and others along the x direction (but we aren't told 
the results). At time t2 > t1 their spin is measured again, this time along the 
x direction, and those with spin up (along x) are saved as a subensemble (those 
with spin down are discarded). Question: Of those remaining (the subensemble), 
what fraction had spin up (along z or x, depending on which was measured) in the 
first measurement?"

\section{Hints example to the sample question}
The equations in the hints are fed as plain latex language. Here they are presented as readable format.

\paragraph{Insightful\_hint:} If $S_z$ was measured at $t_1$, everything was spin up; if $S_x$
was measured, half of them were spin down, but the second measurement did 
nothing, and these were thrown away.

\paragraph{Vague\_hint:} Show that the probability of getting spin up (along $b$) in the 
first measurement is $P_+ = \cos^2(\theta_{ab}/2)$, and (by extension) the 
probability of getting spin up in both measurements is
$P_{++} = \cos^2(\theta_{ab}/2)\cos^2(\theta_{bc}/2)$. Find the other three 
probabilities ($P_{+-}$, $P_{-+}$, and $P_{--}$). Beware: If the
outcome of the first measurement was spin down, the relevant angle is now the 
supplement of $\theta_{bc}$.

\paragraph{Unrelated\_hint:} Example 10.5\\
Yukawa scattering. The Yukawa potential (which
is a crude model for the binding force in an atomic nucleus) has the form
\[
V(r) = \beta \frac{e^{-\mu r}}{r}
\]
where $\beta$ and $\mu$ are 
constants. The Born approximation gives
\[
f(\theta) \approx -\frac{2 m \beta}{\hbar^2 \kappa} \int_0^{\infty} 
e^{-\mu r} \sin (\kappa r) \, dr = -\frac{2 m \beta}{\hbar^2(\mu^2 + 
\kappa^2)} .
\]
Example 10.6\\
Rutherford scattering. If we put in $\beta = q_1 q_2 / 4 \pi \epsilon_0, \mu = 0$, the Yukawa potential reduces to the
Coulomb potential, describing the electrical interaction of two-point charges. 
Evidently, the scattering amplitude is
\[
f(\theta) \approx -\frac{2 m q_1 
q_2}{4 \pi \epsilon_0 \hbar^2 \kappa^2}
\]
or (using Equations 10.89 and 
10.51):
\[
f(\theta) \approx -\frac{q_1 q_2}{16 \pi \epsilon_0 E \sin ^2(
\theta / 2)} .
\]
The differential cross-section is the square of this:
\[
\frac{d \sigma}{d \Omega} = \left[\frac{q_1 q_2}{16 \pi \epsilon_0 E \sin 
^2(\theta / 2)}\right]^2,
\]
which is precisely the Rutherford formula 
(Equation 10.11). It happens that for the Coulomb potential classical mechanics, 
the Born approximation, and quantum field theory all yield the same result. As 
they say in the computer business, the Rutherford formula is amazingly "robust."

\section{Summarization example}

\paragraph{Summarized\_Insightful\_hint:} If \( S_z \) was measured at \( t_1 \), everything was spin up. 
If \( S_x \) was measured, half of them were spin down. But the second measurement did 
nothing, and these were thrown away.

\paragraph{Summarized\_Vague\_hint:} Show that the probability of getting spin up (along \( b \)) 
in the first measurement is \( P_+ = \cos^2(\theta_{ab}/2) \), and (by extension) the 
probabilities of getting spin up in both measurements is \( P_{++} = \cos^2(\theta_{ab}/2) 
\cos^2(\theta_{bc}/2) \). Find the other three probabilities (\( P_{+-} \), \( P_{-+} \), 
and \( P_{--} \)).

\paragraph{Summarized\_Unrelated\_hint:} The Yukawa potential (which is a crude model for the 
binding force in an atomic nucleus) has the form. The Rutherford formula is 
amazingly "robust." It happens that for the Coulomb potential, classical mechanics, 
the Born approximation, and quantum field theory all yield the same result.

\section{Generated Answer}
The equations in the responses are in plain latex language. Here they are presented in readable format.

\paragraph{No\_hint:} All the spins are prepared in the up state along the z axis at \( t = 0 \). This means that initially the state of any spin is \( |up\_z\rangle \). The measurement at \( t_1 \) doesn't change the state, it simply gives us information about the state; however, any given spin is now in a superposition of spin up and spin down along the x direction. The key result you need here is that the \( |up\_z\rangle \) state is a superposition of \( |up\_x\rangle \) and \( |down\_x\rangle \) states. Specifically, 
\[ |up\_z\rangle = \frac{1}{\sqrt{2}} ( |up\_x\rangle + |down\_x\rangle ) \]
Furthermore, the square of the amplitude gives the probability. Thus, prior to the first measurement, each spin has a \( (1/\sqrt{2})^2 = 1/2 \) chance of being measured as \( |up\_x\rangle \), and the same chance of being measured as \( |down\_x\rangle \). When the second measurement is made at time \( t_2 \), only the spins that give \( |up\_x\rangle \) are kept. Since the states \( |up\_x\rangle \) and \( |down\_x\rangle \) were equally likely at \( t_1 \), this discards half the spins. Thus, the fraction of spins remaining that were spin up at \( t_1 \) (either along z or x, whichever was measured) is 1/2. However, keep in mind that the spins measured along the z direction at \( t_1 \) are still in the state \( |up\_z\rangle \), regardless of the later measurement. So, for those spins that were measured along the z direction at \( t_1 \), all of them were spin up. This is independent of the later measurement along the x axis at \( t_2 \). In summary, the final answer will depend on how many spins were measured along x or z at \( t_1 \). But for any given spin, the fraction that were spin up at \( t_1 \) is 1 if measured along z, and 1/2 if measured along x.

\paragraph{Insightful\_hint:} Let's divide the ensemble into two parts based on the measurements at time \( t_1 \): 
1. Those where the spin was measured along the \( z \) direction. 
2. Those where the spin was measured along the \( x \) direction. 
For particles in the first category, since all particles are initially in spin up state with respect to the \( z \) axis, every spin measured will turn out to be up. Therefore, all particles in this category would have been spin up in the first measurement at \( t_1 \). For the second category, where the spin is measured along the \( x \) axis at \( t_1 \), the outcomes will be in a superposition of spin up and spin down along the x axis, due to the angle between the \( z \) and \( x \) direction. When measurement along \( x \) is made, it will have equal probabilities being found in the plus or minus state, due to the orthogonality of the two axes. Therefore, the fraction of particles that would have been measured as spin up during the first measurement for the particles remaining in the subensemble would be: 
- 100
- 50
Without the proportions of the particles that were measured along \( x \) and \( z \), it's not possible to provide a concrete combined percentage. But for measurements along \( z \) it would always be 100

\paragraph{Vague\_hint:} The problem involves sequential measurements of spin 1/2 particles, where the measurements are performed along different axes. Consider the measurement of these particles along the x-axis at time \( t_1 \) (these constitute one part of the original ensemble) and the z-axis at \( t_2 \). The basis states with regard to the z-axis are \( |+z\rangle \) and \( |-z\rangle \) which, when expressed in the x-basis, would be represented as follows: 
1. For spins up along the z-axis: \( |+z\rangle = \frac{1}{\sqrt{2}} ( |+x\rangle + |-x\rangle ) \)
2. For spins down along the z-axis: \( |-z\rangle = \frac{1}{\sqrt{2}} ( |+x\rangle - |-x\rangle ) \)
If we only save the post-measurement states which are spin-up along x, then half of the spins that were measured along x at \( t_1 \) would be thrown away. Therefore, it is given that states of particles can either be spin-up along z or spin-up along x (depending on which was measured) in the first measurement. From here, considering the populations of these particles is consistent with both measurements, we find that: 
1. For the particles that were measured along the z-axis at \( t_1 \), the fraction that remains in the saved subensemble is \( P_{+z} = 1 \).
2. For the particles that were measured along the x-axis, half of the spins that were up along x will remain in the saved subensemble. Thus, the fraction that remains is \( P_{+x} = 0.5 \)
It is important to highlight that these fractions apply under the assumption of ideal measurements. In practice, the actual probabilities could be influenced by experimental uncertainties and possible interactions between particles.

\textbf{Unrelated\_hint:} We don't have concrete numbers for the exact measurements, but let's denote the fractions in general. Let's consider the part of the ensemble on which spin along \( z \) was measured first. The spin was initially prepared in the spin-up state along \( z \), so all these particles had spin up along \( z \) at the first measurement. After the second measurement along \( x \), about 50\% spins will be found up along \( x \) (because spins along \( z \) and \( x \) are not correlated), so this sub-ensemble contributes 50\% with spin up in the first measurement. Now consider the part of the ensemble on which spin along \( x \) was measured first. Again, about 50\% of the spins will be found up along \( x \) in the first measurement, and the projection postulate tells us that these spins are all in the spin-up state along \( x \) afterwards, assuming ideal measurements. When measured again along \( x \), these spins will all still be found up. So this second sub-ensemble also contributes 50\% with spin up in the first measurement. In total, considering these two sub-ensembles combined, out of all particles that are found to have spin up in the second measurement about 50\% (on average) also had spin up in the first measurement (either along \( z \) or along \( x \), depending on the measurement). Remember that these arguments only make sense for large ensembles, as the results for individual particles are stochastic.

\section{Model Answer}

"Answer": "All of them"

\section{Autograde Response}

"Manual": 6

"Auto-1": 7

"Auto-2": 7

\end{document}